\documentclass{article}

\usepackage[accepted]{benelearn2017}

\usepackage[pdftex]{graphicx} 
\usepackage{subfigure}

\usepackage[utf8]{inputenc} 
\usepackage[T1]{fontenc}    
\usepackage{url}            
\usepackage{booktabs}       
\usepackage{amsfonts}       
\usepackage{nicefrac}       
\usepackage{microtype}      

\usepackage{amsmath}
\usepackage{amsthm}
\usepackage{amsfonts}

\usepackage{hyperref}       

\newcommand{\mb}[1]{\mathbf{#1}}

\usepackage{mlapa}

\usepackage{url}

\benelearntitlerunning{Improving Variational Auto-Encoders using convex combination linear Inverse Autoregressive Flow}

\begin{document}

\twocolumn[
\benelearntitle{Improving Variational Auto-Encoders using convex combination linear Inverse Autoregressive Flow}

 \benelearnauthor{Jakub M. Tomczak}{J.M.Tomczak@uva.nl}
 \benelearnauthor{Max Welling}{M.Welling@uva.nl}
 \benelearnaddress{University of Amsterdam, the Netherlands}
\benelearnaddress{{\bf Keywords}: Variational Inference, Deep Learning, Normalizing Flow, Generative Modeling
          }
\vskip 0.3in
]

\begin{abstract}
In this paper, we propose a new volume-preserving flow and show that it performs similarly to the linear general normalizing flow. The idea is to enrich a linear Inverse Autoregressive Flow by introducing multiple lower-triangular matrices with ones on the diagonal and combining them using a convex combination. In the experimental studies on MNIST and Histopathology data we show that the proposed approach outperforms other volume-preserving flows and is competitive with current state-of-the-art linear normalizing flow.
\end{abstract}

\section{Variational Auto-Encoders and Normalizing Flows}

Let $\mb{x}$ be a vector of $D$ observable variables, $\mb{z} \in \mathbb{R}^{M}$ a vector of stochastic latent variables and let $p(\mb{x}, \mb{z})$ be a parametric model of the joint distribution. Given data $\mb{X} = \{\mb{x}_1, \ldots, \mb{x}_N\}$ we typically aim at maximizing the marginal log-likelihood, $\ln p(\mb{X}) = \sum_{i=1}^{N} \ln p(\mb{x}_{i})$, with respect to parameters. However, when the model is parameterized by a neural network (NN), the optimization could be difficult due to the intractability of the marginal likelihood. A possible manner of overcoming this issue is to apply \textit{variational inference} and optimize the following lower bound:
\begin{equation}\label{eq:elbo}
\ln p(\mb{x}) \geq \mathbb{E}_{q(\mb{z}|\mb{x})}[ \ln p(\mb{x}|\mb{z}) ] - \mathrm{KL} \big{(} q(\mb{z}|\mb{x}) || p(\mb{z}) \big{)},
\end{equation}
where $q(\mb{z}|\mb{x})$ is the \textit{inference model} (an \textit{encoder}), $p(\mb{x}|\mb{z})$ is called a \textit{decoder} and $p(\mb{z}) = \mathcal{N}(\mb{z}|\mb{0}, \mb{I})$ is the \textit{prior}. There are various ways of optimizing this lower bound but for continuous $\mb{z}$ this could be done efficiently through a re-parameterization of $q(\mb{z}|\mb{x})$ \cite{KW:13}, \cite{RMW:14}, which yields a \textit{variational auto-encoder} architecture (VAE).

Typically, a diagonal covariance matrix of the encoder is assumed, \textit{i.e.}, $q(\mb{z}|\mb{x}) = \mathcal{N}\big{(} \mb{z}|\boldsymbol\mu (\mb{x}), \mathrm{diag}(\boldsymbol\sigma ^{2}(\mb{x})) \big{)}$, where $\boldsymbol\mu (\mb{x})$ and $\boldsymbol\sigma ^{2}(\mb{x})$ are parameterized by the NN. However, this assumption can be insufficient and not flexible enough to match the true posterior.

A manner of enriching the variational posterior is to apply a \textit{normalizing flow} \cite{TT:13}, \cite{TV:10}. A (finite) normalizing flow is a powerful framework for building flexible posterior distribution by starting with an initial random variable with a simple distribution for generating $\mb{z}^{(0)}$ and then applying a series of invertible transformations $\mb{f}^{(t)}$, for $t=1,\ldots , T$. As a result, the last iteration gives a random variable $\mb{z}^{(T)}$ that has a more flexible distribution. Once we choose transformations $\mb{f}^{(t)}$ for which the Jacobian-determinant can be computed, we aim at optimizing the following lower bound \cite{RM:15} :
\begin{align}\label{eq:nfobjective}
\ln p(\mb{x}) \geq& \mathbb{E}_{q(\mb{z}^{(0)}|\mb{x})} \Big{[} \ln p(\mb{x}|\mb{z}^{(T)}) + \sum_{t=1}^{T} \ln \Big{|}\mathrm{det}\frac{\partial \mb{f}^{(t)} }{ \partial \mb{z}^{(t-1)} } \Big{|} \Big{]} \notag \\
 &- \mathrm{KL} \big{(} q(\mb{z}^{(0)}|\mb{x}) || p(\mb{z}^{(T)}) \big{)}.
\end{align}

The fashion the Jacobian-determinant is handled determines whether we deal with \textit{general normalizing flows} or \textit{volume-preserving flows}. The general normalizing flows aim at formulating the flow for which the Jacobian-determinant is relatively easy to compute. On the contrary, the volume-preserving flows design series of transformations such that the Jacobian-determinant equals $1$ while still it allows to obtain flexible posterior distributions.

In this paper, we propose a new volume-preserving flow and show that it performs similarly to the linear general normalizing flow.

\section{New Volume-Preserving Flow}

In general, we can obtain more flexible variational posterior if we model a full-covariance matrix using a linear transformation, namely, $\mb{z}^{(1)} = \mb{L} \mb{z}^{(0)}$. However, in order to take advantage of the volume-preserving flow, the Jacobian-determinant of $\mb{L}$ must be $1$. This could be accomplished in different ways, \textit{e.g.}, $\mb{L}$ is orthogonal matrix or it is the lower-triangular matrix with ones on the diagonal. The former idea was employed by the Hauseholder flow (HF) \cite{TW:16} and the latter one by the linear Inverse Autoregressive Flow (LinIAF) \cite{KSJCSW:16}. In both cases, the encoder outputs an additional set of variables that are further used to calculate $\mb{L}$. In the case of the LinIAF, the lower triangular matrix with ones on the diagonal is given by the NN explicitly.

However, in the LinIAF a single matrix $\mb{L}$ could not fully represent variations in data. In order to alleviate this issue we propose to consider $K$ such matrices, $\{\mb{L}_{1}(\mb{x}), \ldots, \mb{L}_{K}(\mb{x})\}$. Further, to obtain the volume-preserving flow, we propose to use a convex combination of these matrices $\sum_{k=1}^{K} y_{k}(\mb{x}) \mb{L}_{k}(\mb{x})$, where $\mb{y}(\mb{x}) = [y_1(\mb{x}), \ldots, y_{K}(\mb{x})]^{\top}$ is calculated using the softmax function, namely, $\mb{y}(\mb{x}) = \mathrm{softmax}(\mathrm{NN}(\mb{x}))$, where $\mathrm{NN}(\mb{x})$ is the neural network used in the encoder.

Eventually, we have the following linear transformation with the convex combination of the lower-triangular matrices with ones on the diagonal:
\begin{equation}\label{eq:ccLinIAF}
\mb{z}^{(1)} = \Big{(} \sum_{k=1}^{K} y_{k}(\mb{x}) \mb{L}_{k}(\mb{x}) \Big{)} \mb{z}^{(0)}.
\end{equation}

The convex combination of lower-triangular matrices with ones on the diagonal results again in the lower-triangular matrix with ones on the diagonal, thus, $|\det \Big{(} \sum_{k=1}^{K} y_{k}(\mb{x}) \mb{L}_{k}(\mb{x}) \Big{)} | = 1$. This formulates the volume-preserving flow we refer to as \textit{convex combination linear IAF} (ccLinIAF).

\section{Experiments}

\paragraph{Datasets} In the experiments we use two datasets: the MNIST dataset\footnote{We used the static binary dataset as in \cite{LM:11}.} \cite{MNIST} and the Histopathology dataset \cite{TW:16}. The first dataset contains $28\times 28$ images of handwritten digits (50,000 training images, 10,000 validation images and 10,000 test images) and the second one contains $28\times 28$ gray-scaled image patches of histopathology scans (6,800 training images, 2,000 validation images and 2,000 test images). For both datasets we used a separate validation set for hyper-parameters tuning.

\paragraph{Set-up} In both experiments we trained the VAE with $40$ stochastic hidden units, and the encoder and the decoder were parameterized with two-layered neural networks ($300$ hidden units per layer) and the gate activation function \cite{DG:15}, \cite{DFAG:16}, \cite{OKEVGK:16}, \cite{TW:16}. The number of combined matrices was determined using the validation set and taking more than $5$ matrices resulted in no performance improvement. For training we utilized ADAM \cite{KB:14} with the mini-batch size equal $100$ and one example for estimating the expected value. The learning rate was set according to the validation set. The maximum number of epochs was $5000$ and early-stopping with a look-ahead of $100$ epochs was applied. We used the \textit{warm-up} \cite{BVVDJB:15}, \cite{SRMSW:16} for first $200$ epochs. We initialized weights according to \cite{GB:10}.

We compared our approach to linear normalizing flow (VAE+NF) \cite{RM:15}, and finite volume-preserving flows: NICE (VAE+NICE) \cite{DKB:14}, HVI (VAE+HVI) \cite{SKW:15}, HF (VAE+HF) \cite{TW:16}, linear IAF (VAE+LinIAF) \cite{KSJCSW:16} on the MNIST data, and to VAE+HF on the Histopathology data. The methods were compared according to the lower bound of marginal log-likelihood measured on the test set.

\begin{table}[t]
\caption{Comparison of the lower bound of marginal log-likelihood measured in nats of the digits in the MNIST test set. Lower value is better. Some results are presented after: $\clubsuit$ \cite{RM:15}, $\diamondsuit$ \cite{DKB:14}, $\spadesuit$ \cite{SKW:15}.}
\label{tab:mnist}
\vskip 0.15in
\begin{center}
\begin{small}
\begin{sc}
\begin{tabular}{ll}
    Method & $\approx \ln p(\mb{x})$ \\
    \hline
    VAE & $-93.9$ \\
    \hline
    VAE+NF ($T$=10) $\clubsuit$	& $-87.5$ \\    
    VAE+NF ($T$=80) $\clubsuit$	& $-85.1$ \\
    \hline
	VAE+NICE ($T$=10) $\diamondsuit$ & $-88.6$ \\
	VAE+NICE ($T$=80) $\diamondsuit$	& $-87.2$ \\
	VAE+HVI ($T$=1) $\spadesuit$	& $-91.7$ \\
	VAE+HVI ($T$=8) $\spadesuit$	& $-88.3$ \\
    VAE+HF($T$=1) & $-88.1$ \\
    VAE+HF($T$=10) & $-87.8$ \\
    \hline
    VAE+LinIAF & $-86.7$ \\
    VAE+ccLinIAF($K$=5) & $-86.1$ \\
\end{tabular}
\end{sc}
\end{small}
\end{center}
\vskip -0.1in
\end{table}

\begin{table}[t]
\caption{Comparison of the lower bound of marginal log-likelihood measured in nats of the image patches in the Histopathology test set. Higher value is better. The experiment was repeated $3$ times. The results for VAE+HF are taken from: $\heartsuit$ \cite{TW:16}.}
\label{tab:histopathology}
\vskip 0.15in
\begin{center}
\begin{small}
\begin{sc}
\begin{tabular}{ll}
    Method & $\leq \ln p(\mb{x})$ \\
    \hline
    VAE $\heartsuit$ & $1371.4 \pm 32.1$ \\
    VAE+HF ($T$=1) $\heartsuit$  & $1388.0 \pm 22.1$ \\
    VAE+HF ($T$=10) $\heartsuit$ & $1397.0 \pm 15.2$ \\
    VAE+HF ($T$=20) $\heartsuit$ & $1398.3 \pm 8.1$ \\
    \hline
    VAE+LinIAF & $1388.6 \pm 71$ \\
    VAE+ccLinIAF($K$=5) & $\mb{1413.8 \pm 22.9}$
	\end{tabular}
\end{sc}
\end{small}
\end{center}
\vskip -0.1in
\end{table}

\paragraph{Discussion} The results presented in Table \ref{tab:mnist} and \ref{tab:histopathology} for MNIST and Histopathology data, respectively, reveal that the proposed flow outperforms all volume-preserving flows and performs similarly to the linear normalizing flow with large number of transformations. The advantage of using several matrices instead of one is especially apparent on the Histopathology data where the VAE+ccLinIAF performed better by about $15$nats than the VAE+LinIAF. Hence, the convex combination of the lower-triangular matrices with ones on the diagonal seems to allow to better reflect the data with small additional computational burden.

\paragraph{Implementation} The code for the proposed approach can be found at: \url{https://github.com/jmtomczak/vae_vpflows}.

\section*{Acknowledgments}
The research conducted by Jakub M. Tomczak was funded by the European Commission within the Marie Sk\l odowska-Curie Individual Fellowship (Grant No. 702666, ''Deep learning and Bayesian inference for medical imaging'').

\bibliographystyle{mlapa}

\end{document}